\documentclass[letterpaper, 10 pt, conference]{ieeeconf}  

\IEEEoverridecommandlockouts                              

\usepackage{graphicx}      
\usepackage{algorithm} 
\usepackage{algorithmic}
\usepackage{varwidth} 
\usepackage{gensymb}
\usepackage{amsfonts}
\usepackage{cite}
\usepackage{amsmath}
\usepackage{url}
\usepackage{hyperref}
\usepackage{multicol}
\usepackage[caption=false,font=footnotesize]{subfig}

\graphicspath{{figures/}}
\title{\LARGE \bf
6-DoF Contrastive Grasp Proposal Network
}

\author{Xinghao Zhu$^{*,1}$, Lingfeng Sun$^{*,1}$, Yongxiang Fan$^{2}$ and Masayoshi Tomizuka$^{1}$
\thanks{*The authors contributed equally to this work}
\thanks{$^1$Department of Mechanical Engineering, 
        University of California, Berkeley, Berkeley, CA 94720, USA.
        {\tt\small {zhuxh, lingfengsun, tomizuka}@berkeley.edu}}
\thanks{$^2$FANUC Advanced Research Laboratory, FANUC America Corporation, Union City, CA, USA.
{\tt\small {Yongxiang.Fan@fanucamerica.com}}
}}

\begin{document}

\maketitle

\thispagestyle{empty}
\pagestyle{empty}

\begin{abstract}
Proposing grasp poses for novel objects is an essential component for any robot manipulation task. Planning six degrees of freedom (DoF) grasps with a single camera, however, is challenging due to the complex object shape, incomplete object information, and sensor noise. In this paper, we present a 6-DoF contrastive grasp proposal network (CGPN) to infer 6-DoF grasps from a single-view depth image. First, an image encoder is used to extract the feature map from the input depth image, after which 3-DoF grasp regions are proposed from the feature map with a rotated region proposal network. Feature vectors that within the proposed grasp regions are then extracted and refined to 6-DoF grasps. The proposed model is trained offline with synthetic grasp data. To improve the robustness in reality and bridge the simulation-to-real gap, we further introduce a contrastive learning module and variant image processing techniques during the training. CGPN can locate collision-free grasps of an object using a single-view depth image within 0.5 second. Experiments on a physical robot further demonstrate the effectiveness of the algorithm. The experimental videos are available at~\cite{website}.
\end{abstract}

\section{Introduction}
\label{intro}
Robotic grasping in unstructured environments can benefit applications in manufacturing, retail, service, and warehousing. 
Grasping unseen objects is, however, highly challenging due to the limitations in perceptions. 
When objects are cluttered in a bin, exact geometry and position of objects are obscured. Sensing imprecision and deficiency then leads to poor grasp planning execution.

Model-based and learning-based methods could be used to plan grasps across a wide variety of objects. 
Existing physical grasp analysis techniques, such as grasp quality metrics~\cite{quality_metrics}, template matching~\cite{template_matching}, and wrench space analysis~\cite{206book}, can be used to search for the optimal grasp. These approaches, however, can be less robust in practice due to the perception limitation. Incompletion of the object surface can lead to flawed analysis.
An alternative approach is to plan grasps with supervised deep learning. 
Current methods show that it is preferable to learn grasp quality functions and optimize them at the runtime~\cite{dexnet2, dexnet3, 6dofgraspnet, 6dofgraspnet_inclutter, gpd, ppojpo, finger_splitting, surface_fitting, pointgpd}.
Learning intermediary information, such as grasp qualities and success rate, can improve the training efficiency and prediction accuracy. However, the requirement of the sampling or optimization makes the algorithm time-consuming.
To tackle this, other methods use end-to-end learning to infer grasp poses from the sensor inputs directly~\cite{pointnetgrasp,unigrasp}. Nevertheless, these algorithms require larger datasets and elaborate hyper-parameters tunning to reduce the training variance.

\begin{figure}[tb]
\begin{center}
	\includegraphics[width=2.6in]{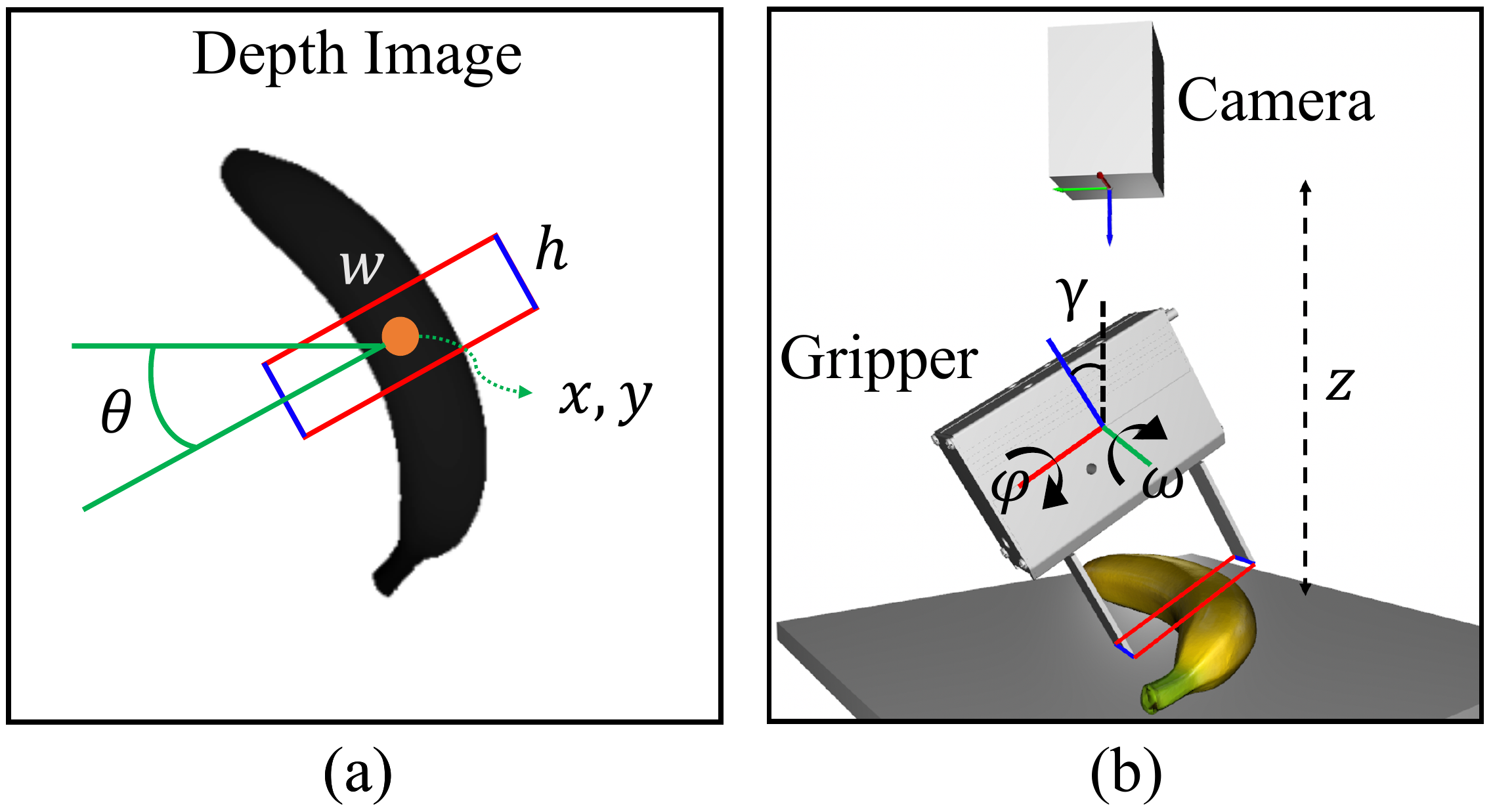}
	\caption{Grasp Representation $(x, y, \theta, \gamma, z, \beta)$. The planar 3D grasp pose $(x, y, \theta)$ in (a) represents the center position and orientation of the projected bounding box on camera plane. The bounding box's width $w$ and height $h$ are used in training but not in representing grasps.
	The other 3 grasp parameters are shown in (b): tilt angle $\gamma$ is the rotation among axis $\omega$, $z$ is the depth of grasp, and gripper angle $\beta$  is the rotation among the grasp axis $\varphi$.}
	\label{fig: grasp}
\end{center}
\end{figure}

Ideally, the grasp planning algorithm generates 6 degrees of freedom (DoF) grasps. Previous works have proposed models that can detect top-down grasps using depth images~\cite{dexnet2,rpn_grasp_2,rpn_grasp_3}. However, 
the top-down nature of such grasps 
does not allow robots to pick up objects from different orientations, which limits its application in cluttered environments. 
In this work, we propose to use six variables $(x,y,\theta,\gamma,z,\beta)$ to represent a 6-dimensional grasp in a depth image, as shown in Fig.~\ref{fig: grasp}. 
Three  planar 3D grasp poses $(x, y, \theta)$ in Fig.~\ref{fig: grasp}(a) are the center position and orientation of the bounding box in the image plane. The bounding box's width $w$ and height $h$ are used in training but not in representing grasps.
Three spatial grasp parameters shown in Fig.~\ref{fig: grasp}(b) are: the tilt angle $\gamma$ among axis $\omega$, the rotation angle $\beta$ among the grasp axis $\varphi$, and the depth of the grasp $z$. 

In previous works~\cite{dexnet2,dexnet3,6dofgraspnet,6dofgraspnet_inclutter}, the strategy to train on synthetic datasets and apply to reality has been heavily used. It has been shown that rendered images with numerically computed grasp qualities can ease the data preparation process.
The simulation-to-real (sim-to-real) gap, however, is still an open problem for grasp planning. Synthetic data have better resolution and less noise than real images. 
To tackle this problem, we introduce contrastive learning with sim-to-real depth image processing in this paper. 
Contrastive learning aims to extract invariant features from augmented images, which improves the overall performance of modeling under vision noise.

\begin{figure}[tb]
\begin{center}
	\includegraphics[width=3.4in]{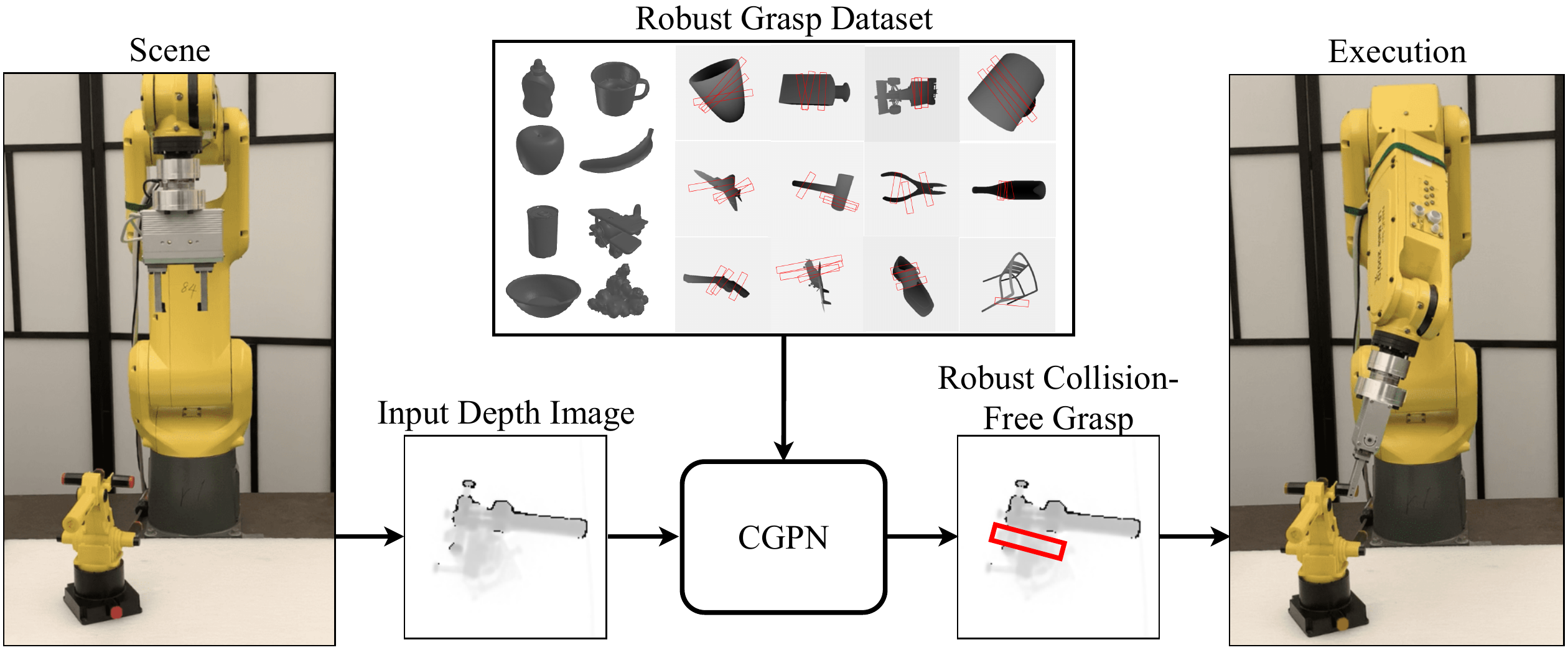}
	\caption{CGPN Architecture. The 6-DoF contrastive grasp proposal network (CGPN) is trained offline to infer grasps from depth images using a dataset of synthetic images and grasps. When an object is presented to the robot, a stereo camera captures a depth image; CGPN could rapidly generate 6-DoF robust collision-free grasps, which is executed with the Fanuc robot.}
	\label{fig: framework}
\end{center}
\end{figure}

In this paper, we propose an end-to-end 6-DoF contrastive grasp proposal network (CGPN). The general framework is shown in Fig~\ref{fig: framework}.
When an object is presented in the scene, a stereo camera captures a depth image; CGPN rapidly generates 6-DoF robust grasps, which are executed with the Fanuc robot.
CGPN is trained with synthetic grasp images with variant augmentation techniques to bridge the sim-to-real gap.

The contributions of this paper are as follows.
\begin{enumerate}
  \item An end-to-end grasp planning model is proposed to detect grasps efficiently. The model consists of a feature encoder, a rotated region proposal network, a grasp refinement network, and a collision detection module. The model uses single-view depth images as input and infers 6-DoF grasps for parallel-jaw grippers.
  \item A contrastive learning module and a depth image processing technique are introduced in the grasp planning to resolve the sim-to-real gap.
\end{enumerate}

The remainder of this paper is organized as follows.
Related works are introduced in Section~\ref{sec: related_work}.
Section~\ref{sec: algorithm} presents the proposed grasp planning algorithms in detail.
Experiments are presented in Section~\ref{sec: exp}.
Section~\ref{sec: conclusion} concludes the paper and introduces the future work.

\section{Related Works}
\label{sec: related_work}

\subsection{Grasping}
Recent learning-based approaches have demonstrated promising efficiency and robustness in the grasp planning task.
Some works propose to optimize learned grasp quality metrics at runtime~\cite{dexnet2, 6dofgraspnet, 6dofgraspnet_inclutter, gpd, pointgpd, surface_fitting}. 
Sampling grasp candidates among the object's surface, however, can be time-consuming and requires pre-defined heuristics.
Other approaches~\cite{pointnetgrasp, unigrasp, roi_grasp,rpn_grasp_2,rpn_grasp, rpn_grasp_3, rpn_grasp_4} propose to directly infer grasp poses from the raw input with an end-to-end model. 
These approaches design a deep model to perform grasp pose regression and grasp quality assessment at the same time. 

To represent grasps in the input image, rotated rectangle representation, or rotated bounding box, has been utilized by previous work~\cite{grasp_with_rgbd}.
~\cite{grasp_with_rgbd} demonstrates that by using a two-step search strategy, 3-DoF grasps could be located within the image plane. This algorithm, however, uses comprehensive searching in the whole input image, thus is time-consuming.

\subsection{Region Proposal Network}

The success of the region proposal network (RPN)~\cite{fast_rcnn, faster_rcnn} brings focus to the bounding box grasp representation. 
The objective of RPN is to classify and regress object's bounding boxes in images.
Instead of sampling candidates in the original image plane, RPN utilizes shared feature maps. Classification and regression are performed in the latent feature spaces, which has been shown to have promising accuracy and time complexity. \cite{rpn_grasp, roi_grasp, rpn_grasp_2,rpn_grasp_3,rpn_grasp_4} introduce the RPN network to the grasp planning. These approaches, however, focus on 3-DoF top-down grasps and require an additional grasp detector to predict the rotation angle. 
In this work, we propose to infer rotated grasps directly from the image plane. Besides, we use a downstream grasp refinement network to regress the full 6-DoF grasps.

\subsection{Contrastive learning}
Contrastive learning\cite{Hadsellcontrastive}, and many of its recent applications~\cite{he2020momentum,chen2020simple} in visual representation learning, can be thought of as training an encoder to extract invariant features for similar images. This encoding mechanism helps maintain the understanding of the same object under correlated views. Contrastive losses measure the similarities of sample pairs in representation space. \cite{dosovitskiy2015discriminative} uses the latent parametric feature to represent instance as a class.\cite{zhuang2019local, tian2020contrastive} introduce memory-bank based methods and momentum to store and update the class representation. For grasping, this helps to reduce the instability of the model resulting from the sim-to-real gap and various input noise in a cluttered environment. In our framework, we generate stable grasp proposals by leveraging recent advances in data augmentation and contrastive loss.


\section{6-DoF Contrastive Grasp Proposal Network}
\label{sec: algorithm}
\subsection{General Framework}
\begin{figure*}[t]
\begin{center}
	\includegraphics[width=6.7in]{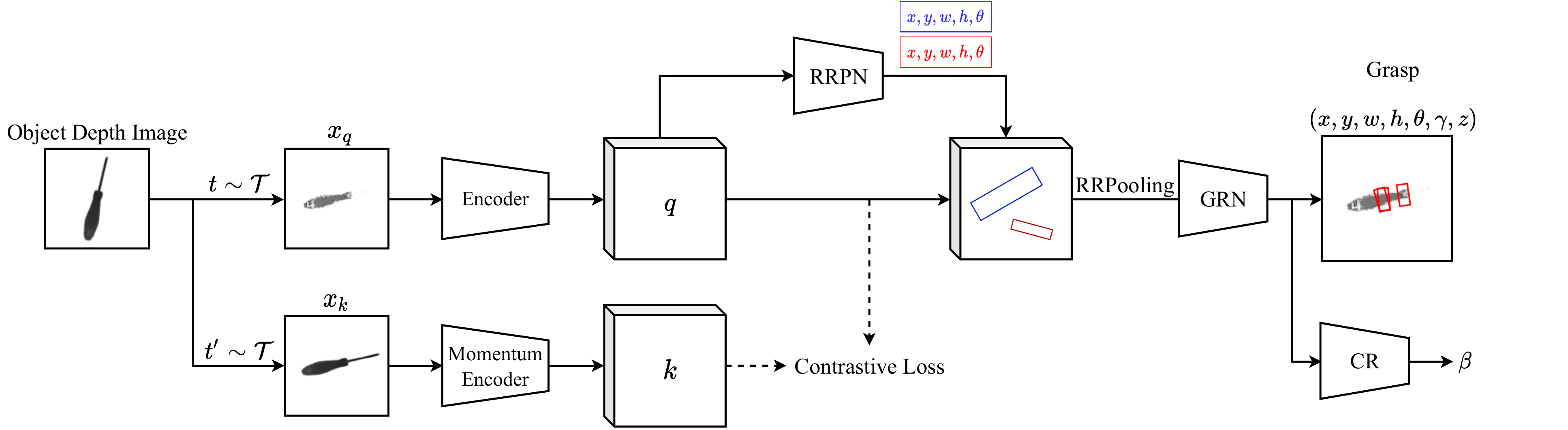}
	\caption{CGPN Network Architecture. 
	During the training, a synthetic depth image of the object is feeding into the network. 
	First, two separate data augmentation operators $t,t'$ are applied to the segmented depth image to obtain $x_q, x_k$, which are then input to the contrastive encoders. 
	Second, extracted feature maps $q,k$ are parsed to the rotated region proposal network (RRPN) to generate grasp regions. 
	Third, a rotated region pooling (RRPooling) module extracts feature vectors from the feature map $q$ using the generated grasp regions.
	Finally, a grasp refinement network (GRN) infers the tilt angle $\gamma$ and the depth $z$ using the local feature vectors. A collision refinement is further added to search the rotation angle $\beta$.}
	\label{fig: network}
\end{center}
\end{figure*}

This section introduces the framework of the 6-DoF contrastive grasp proposal network, as shown in  Fig.~\ref{fig: network}. The input to the whole pipeline is a single-view depth image of the scene. Segmentation is first applied to the image to separate objects. The CGPN model generates grasps for the object using its segmented depth image. 

\subsubsection{Training Phase} During training, the CGPN algorithm works as follows. First, two separate data augmentation operators $t,t'$ are sampled from the augmentation family $\mathcal{T}$. Augmentations are applied to the segmented depth image to obtain two correlated views $x_q, x_k$  regarded as a positive pair in the contrastive learning module. Similar to~\cite{he2020momentum}, the other inputs in the same batch are viewed as negative samples. Second, the positive and negative samples are fed into the contrastive encoder. We get query $q$ from $x_q$ with query encoder and keys $\{k_i\}_{1}^N$ from $x_k$ and other negative samples with key encoder. The key encoder is a slowly updated query encoder with no gradient update.
Contrastive loss is calculated using these queries and keys, with more information introduced in the loss section. The purpose of this module is to maximize the agreement of encoded feature maps $q,k$. We assume the encoder can learn invariant representations of the depth images under different augmentation operations, and the query $q$ is the feature we use for downstream tasks. Third, a rotated region proposal network (RRPN) is introduced to propose 3-DoF grasp regions base on the feature map $q$. The output of RRPN  determines 3-DoF grasp $(x, y, \theta)$ and a box shape $(w, h)$. We still need local features to determine other grasp parameters. Fourth, a rotated region pooling (RRPooling) module extracts feature vectors from the feature map $q$ using the predicted rotated bounding box from RRPN. These local feature maps contain depth and other features around the proposed grasp position. Finally, a grasp refinement network (GRN) is designed to infer the tilt angle $\gamma$ and the depth $z$ using the local feature vectors. This completes a 5-DoF grasp pose together with $(x, y, \theta)$. The last DoF $\beta$ is searched with a collision refinement (CR) module, thus is neglected in the training process.
The training loss is the weighted combination of the contrastive loss, region proposal loss, and the grasp refinement loss. After training, we get a query encoder, a RRPN module, and a GRN module used for online testing.

\subsubsection{Testing Phase}
During the testing phase, we do not need to create positive pairs for the contrastive modules. Instead, we directly put the input depth image into the query encoder. The following process is the same as training. The output of the CGPN model is valid grasp poses for each object. After that, the 5-DoF grasps for each object are projected back to the original cluttered scene. The last DoF $\beta$ is determined according to the collision constraints.

\subsection{Data Augmentation and Contrastive Learning}
The instability in real-world grasping is usually brought by occlusion in cluttered scenes, background noise in the environment, and the sim-to-real gap. To overcome this, we design multiple data augmentation operations $\mathcal{T}$ for the input depth image.

\begin{figure}[htbp]
\begin{center}
	\includegraphics[width=3.4in]{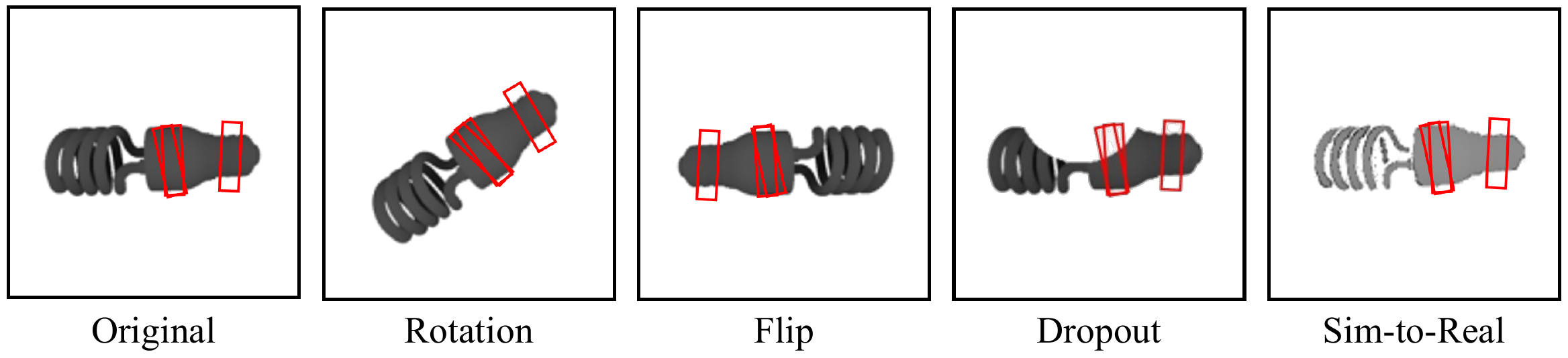}
	\caption{Illustrations of the available data augmentation operators in $\mathcal{T}$. Each augmentation can transform the original input with some internal parameters (e.g. rotation degree, flip axis). We use combination of the first four operations and the sim-to-real process as a complete augmentation for a single image.}
	\label{fig: aug}
\end{center}
\end{figure}

\subsubsection{Spatial/geometric transformation} 
Spatial operations include rotation, flip, and dropout on the original synthetic depth image. Each augmentation can transform the input image stochastically with some internal parameters. We use composition of these operations to create different observations for the same grasp object. Note that unlike classification labels in vision, the ground-truth high-quality grasp poses would change along with these augmentation operations. 
\subsubsection{Sim-to-Real Processing}
Similar to \cite{rope_manipulation}, we design a sim-to-real transfer operation by leveraging several image processing techniques. We randomly paint the pixels black in areas of high Laplacian gradient and edges of the object detected by Canny edge detector\cite{cannyedge}. Then we inject realistic noise into the image. The operation is parameterized by a threshold of black paining areas. Compared to spatial transforms, the sim-to-real process does not change the ground-truth pose in the image. 

Illustrations of the data augmentation operators are shown in Figure \ref{fig: aug}. Using the introduced operations, we extend the training dataset and create positive pairs from the same input depth image for the contrastive training process. The transformation family $\mathcal{T} $ is created by a random combination of the spatial operations and the sim-to-real transfer as the last operation. All of the operations have an execution probability. In this way, we can train on synthetic data generated in the simulator and use the model to predict valid grasp in real-world scenarios.

Regarding the contrastive learning module in the pipeline, as described in previous sections, we use separate encoders for query input $x_q$ and other positive or negative samples as suggested in \cite{he2020momentum}. Note that some operations in the data augmentation change the ground-truth in the downstream grasping tasks (e.g. spatial operations), directly maximizing the similarity between $q$ and $k^+$ might not be suitable for grasping regression. As suggested in~\cite{chen2020simple}, we add a multi-layer perceptron (MLP) after the query and key encoders as a projecting head and get $\hat{q}, \hat{k_i}$. This is neglected in the network architecture but implemented in experiments. The query and key vector $q, k_i$ can keep the differences in grasping, but the projecting head would catch the invariant properties among positive pairs. We use $q$ for downstream grasping tasks, but use $\hat{q}$ and $\hat{k_i}$ for calculating contrastive loss.

\subsection{Rotated Region Proposal}
Another core part of our architecture is the rotated region proposal network. Similar to the architecture in \cite{rrpn}, which output detected bounding box for scene text, our network receives feature maps from the learned contrastive encoder and output candidate region proposals with class labels and parameters of the positions. The class label determines if the proposal fits any robust grasp, and the parameters $(x, y, w, h, \theta)$ gives a rotated bounding box. 

Unlike text detection, we do not have a fixed number of non-overlapping labelled bounding boxes for each input image. For each object and its depth image, the high-quality grasp poses may overlap with others. When matching the ground-truth grasps with proposed regions, overlapped grasps may have similar skew intersection over union (IoU) results. Directly choosing the one with the highest IoU may cause all proposals matching the ``largest" grasp (i.e., large $w, h$). To avoid such mode collapse, we introduce to random select among top-$k$ grasps based on IoU during matching.

For parameters of the proposed region, although only $(x, y, \theta)$ is used in a grasp pose, the box's width $w$ and height $h$ indicate the range of local features used in the following GRN. The GRN model first aligns and extracts rotated regions of interest by projecting proposals from the RRPN onto the feature map and then use the local features to predict the tilt angle and relative depth of the grasp, which finalizes the 5-DoF grasp $(x, y, \theta, \gamma, z)$.

\subsection{Collision Refinement}
The grasp proposal models mentioned above are used for 5-DoF grasps for a single object.
To use the proposed grasps in a cluttered scene, we use collision constraints to infer all 6 DoFs.
The proposed 5-DoF grasp $g=(x, y, \theta, \gamma, z) \in \mathcal{G}$ are frozen and a posterior optimization process is used to search for the last rotation angle $\beta$:
\begin{equation}
\begin{aligned}
\min_{\beta} \quad & 
\sum_{i}^N C(g, \beta, x_i)\\
\end{aligned}
\end{equation}
where $C$ is the collision check score for a 6-DoF grasp $(g,\beta)$ and surrounding objects. $\{x_i\}_{i=1}^N$ is the segmented depth image for the $N$ object in the scene.

Sign-distance field is introduced in this paper to model the collision score:
\begin{equation}
\begin{aligned}
C(g, \beta, x_i) = -SD(FK(g,\beta), x_i)
\end{aligned}
\end{equation}
where $FK(\cdot)$ denotes the forward kinematics function of the robot. $SD(\cdot)$ denotes the signed-distance function of the robot and the object $x_i$.
Given a 5-DoF grasp, there are infinitely many grasp candidates since the rotation among the grasp axis $\varphi$ is free-floating.
By minimizing the negative signed distance between the robot and the object, a unique 6-DoF grasp can be determined.

\subsection{Loss Design}
\subsubsection{Region Proposal Loss}
Despite the randomness we introduced in positive region matching, most of the model we use in RRPN is similar to \cite{rrpn}. The loss function for the proposal takes the form of:
\begin{equation}
    L_p = -\log s_{\mathrm{pos}} + \sum_{v\in \{x, y, \theta, h ,w\}}\lambda_i \mathrm{smooth}_{L_1} (v^*-v)
\end{equation}
where $s_{\mathrm{pos}}$ is the matching IoU for positive pairs, $v^*$ is the ground-truth value for corresponding variable. The smoothed $L_1$ loss is:
\begin{equation}
    \mathrm{smooth}_{L_1}(x)=\begin{cases} 
      0.5x^2 \quad & \mathrm{if}\; |x|<1  \\
      |x|-0.5 & \mathrm{otherwise} \\
   \end{cases}
\end{equation}
The output $(x, y, \theta)$ is grasp parameter and $(w, h)$ are parameters for local feature ranges in original feature map. Therefore we assign more weight on $\lambda_x, \lambda_y$ and $\lambda_{\theta}$ and less on $\lambda_w, \lambda_h$.
\subsubsection{Grasp Refinement Loss}
To find the best grasp pose, we need not only the rotated planar grasp position proposed by RRPN but also the tilt angle $\gamma$ and depth $z$ of grasp. While RRPN gives us a rough estimation of grasp positions, GRN uses the proposed region of interests (ROIs) to generate accurate grasp poses with local features. We formulate a weighted refinement loss in Eq. (\ref{Eq:refine}) to minimize the L1 error of tilt and depth.
\begin{equation}
    L_r=\lambda_{\gamma} \lVert \hat{\gamma} - \gamma \rVert_1 + \lambda_{z} \lVert \hat{z} - z \rVert_1
    \label{Eq:refine}
\end{equation}

\subsubsection{Contrastive Loss}
Unlike regression loss, the contrastive loss does not have supervised signals. For simplicity, we use $q, k_i$ to represent $\hat{q}, \hat{k_i}$ after the projecting head. Consider a query $q$ encoded from sample $x$ and a dictionary of $N$ keys $\{k_1, k_2, ...,k_N\}$ encoded from different samples. Among all keys, there is one positive key $k^+$ encoded from $x_k$ similar to $x_q$ and $N-1$ negative keys from other samples in the batch. Using dot product as the similarity metric, the loss function is defined as Eq. (\ref{Eq:contrastive}), where $\tau$ is a temperature hyper-parameter.

\begin{equation}
    L_q=-\log \frac{\exp(q\cdot k^+)}{\sum_{i=1}^{N}\exp(q\cdot k_i)}
    \label{Eq:contrastive}
\end{equation}
 This is the log loss of a $N$-way softmax-based classifier that tries to classify $q$ as $k^+$, introduced as InfoNCE in \cite{oord2019representation}. 

The overall loss function used to train the contrastive grasp proposal model is:
\begin{equation}
    L_{\mathrm{overall}}=\lambda_p L_p+\lambda_r L_r+ \lambda_q L_q
\end{equation}
We adjust the relative weight of the three losses at different stages of training. In the beginning, we set the loss of RRPN and GRN low to reduce the contrastive loss. After we get a stable encoder, we increase the weight of RRPN loss but keep the GRN loss weight low to train the 3-DoF grasp position and local feature bounding box. We then gradually increase the weight of GRN to train the overall pipeline for the final 5-DoF grasp. The high weight of GRN loss at an early stage would affect RRPN's training since the models work in series. 



\section{Experiment}
\label{sec: exp}

\subsection{Dataset Generation}
\label{subsec: data_gen}
To generate the grasp sets, we need to sample a large number of grasps for single objects and label the top-ranked robust grasps as ground truths. This is unrealistic for real robots but not hard in simulation environments. We train our CGPN model on the generated single object grasp dataset and use it on real robots. Similar to~\cite{dexnet2}, 1,366 objects are selected from the 3DNet~\cite{3dnet} as the object set. 100 antipodal grasps are evenly sampled among the surface for each object. Each grasp is labeled with the robust force closure metric and is represented by its contact points $(c_1, c_2)$ in 3D.
Since the CGPN algorithm requires depth images as input, objects and grasps are projected to the image plane.
For each selected object, 20 synthetic depths images are rendered from different angles. The object is placed at the center of a regular icosahedron; cameras are placed at each face's center and point to the origin.
The distance between the camera and the object is sampled from $\mathcal{U}(\sqrt{3} r_{obj}, 2 r_{obj})$, where $\mathcal{U}$ denotes the uniform distribution and $r_{obj}$ is the object bounding ball's radiance. Such selection makes sure that the full object is visible in the camera.

\begin{figure}[tb]
\begin{center}
	\includegraphics[width=3.4in]{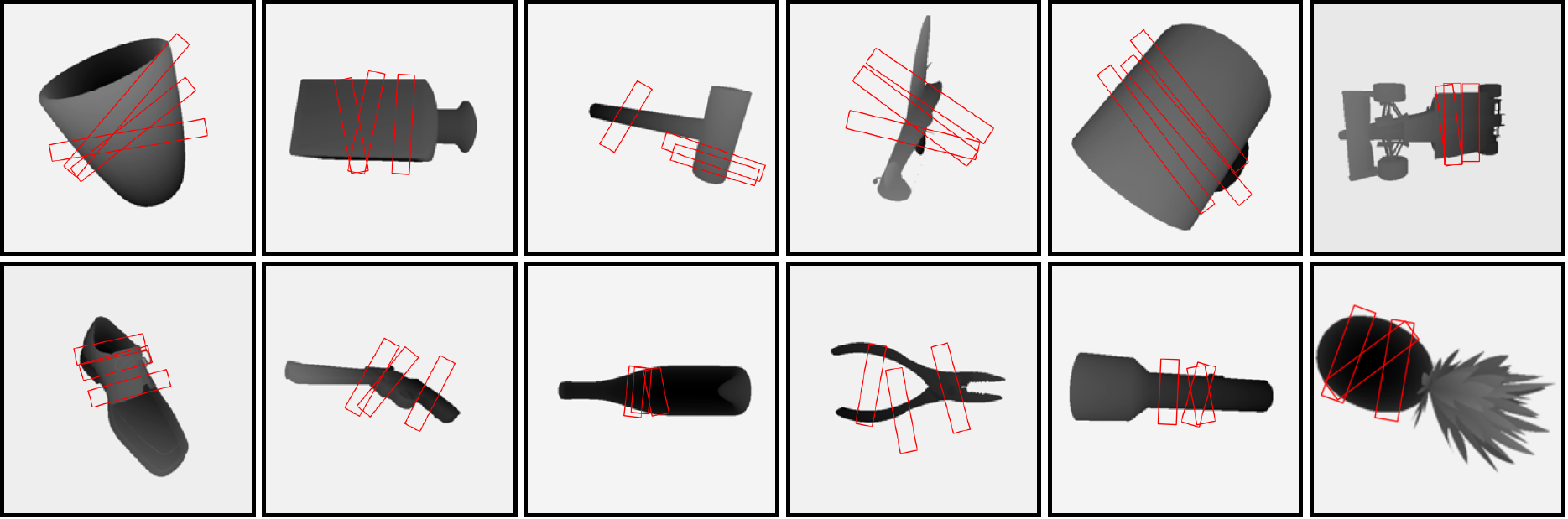}
	\caption{Dataset Samples. This figure shows 12 samples from the generated grasp dataset. 3D grasps are projected to the image plane (red rectangles). The tilt angle $\gamma$ and the distance $z$ are neglected in the plot for simplicity. }
	\label{fig: dataset_samples}
\end{center}
\end{figure}

Each grasp $(c_1, c_2)$ is then projected to the depth image using projective transformations:
\begin{equation}
\begin{bmatrix}
u_i & v_i & f
\end{bmatrix}^T = K \cdot \begin{bmatrix}R & t \end{bmatrix} \cdot
\begin{bmatrix}X_i & Y_i & Z_i & 1 \end{bmatrix}^T
\end{equation}
where $X_i,Y_i,Z_i$ are positions of the contact point $\{c_i\}_{i=1}^2$ in the camera frame, $K , \begin{bmatrix}R & t \end{bmatrix}$ are the camera's intrinsic and extrinsic matrices, respectively. $u_i,v_i$ are pixel's locations at the image for point $c_i$.
The bounding box's other parameters are then computed.
The width $w$ and the height $h$ of the box is set to $\left \| [u_2 - u_1, v_2 - v_1] \right \|_2$ and 20 respectively. Grasp depth $z$, rotation angle $\theta$, and tilt angle $\gamma$ are computed as $z=\frac{1}{2}(Z_1+Z_2)$, $\theta=\tan^{-1} \frac{u_2 - u_1}{v_2 - v_1}$ and $\gamma=\tan^{-1} \frac{Z_2 - Z_1}{w}$ respectively.

Ground truth grasps for each image are selected as the top $20\%$ from 100 samples. 
In this paper, we limit the tilt angle's range to $[-30\degree, 30\degree]$.
To give rotation angle $\theta$ and tilt angle $\gamma$ unique definition, we set constrains on the grasp points in the image plane, such that $v_2 > v_1$. In other words, the point $[u_2, v_2]^T$ is always on the right of the point $[u_1, v_1]^T$. Then, we bound $\theta$ in the range of $[-90\degree, 90\degree]$, and $\gamma$ is relabeled with the corresponding sign. The generated training dataset has 24,723 depth images with labeled ground truth grasps.
Fig.~\ref{fig: dataset_samples} shows some samples from the dataset.

\begin{figure}[tb]
\begin{center}
	\includegraphics[width=3.4in]{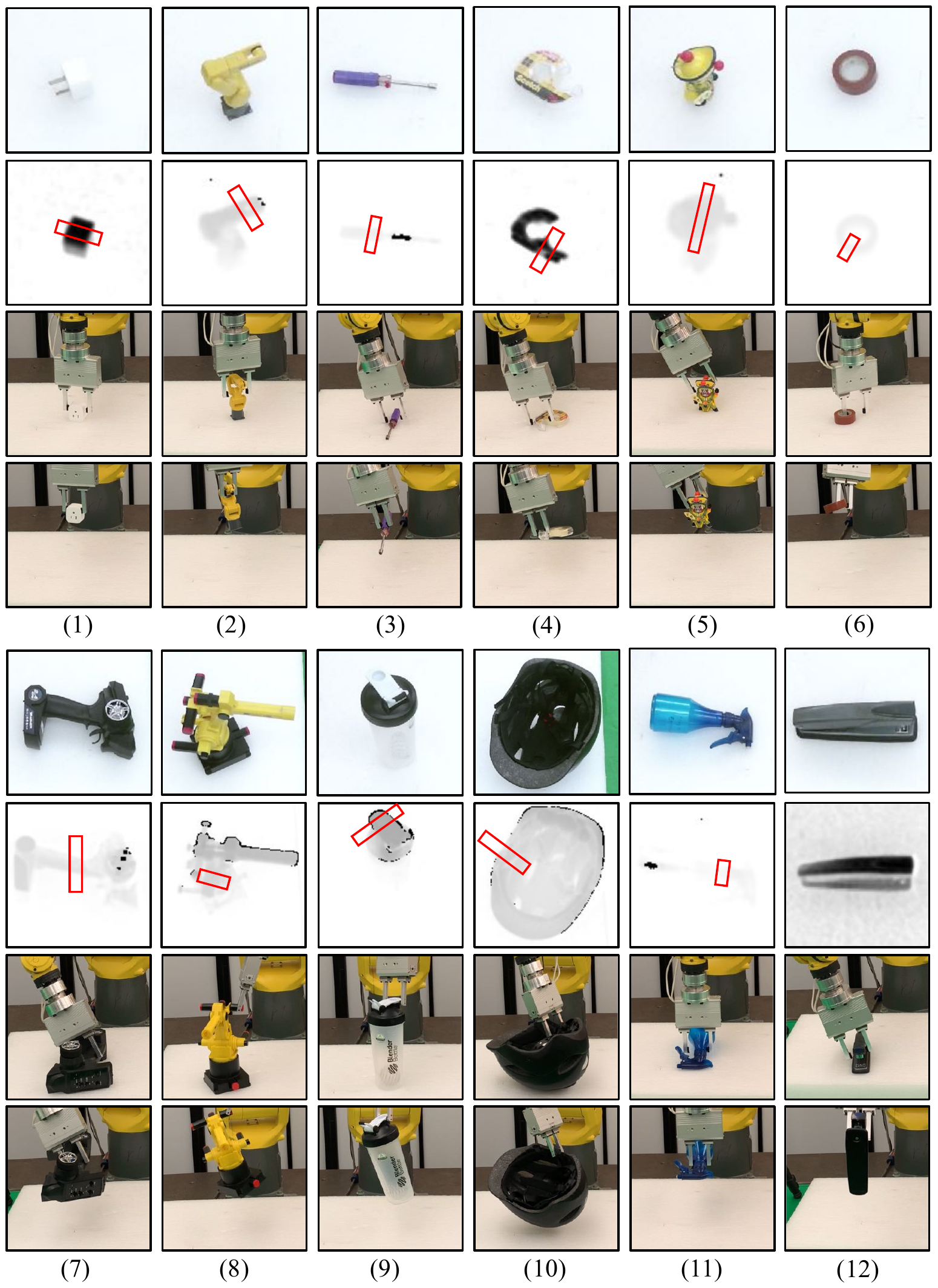}
	\caption{(1-12) The grasp planning and execution results on 12 objects with a single depth image. For each column, (Top Two Rows) show perceived RGB and depth images with planned grasps, (Third Row) shows physical grasps reaching the target grasp, and (Bottom Row) shows the execution results. The tilt angle $\gamma$, and the distance $z$ are neglected in the plot for simplicity. }
	\label{fig: single_results}
\end{center}
\end{figure}

\subsection{Experiment Results}
The proposed CGPN is run on a desktop with GTX2080Ti GPU, 32GB RAM, and 4.0GHz CPU. 
For the experiment, we use a FANUC LRMate 200iD/7L industrial manipulator with a SMC LEHF20K2-48-R36N3D parallel-jaw gripper for grasping. A Kinect v2 camera is used to capture depth image of the scene.
The point cloud library~\cite{pcl} implementation of region growing method is utilized to pre-process and segment the object.

We leverage state-of-the-art model architectures for each submodel.
ResNet-50~\cite{resnet} and rotated region proposal networks~\cite{rrpn} are utilized as the encoder and downstream models.
The hyper-parameter for RRPN and contrastive learning are mostly the same as introduced in \cite{rrpn,he2020momentum} and $\lambda_x=\lambda_y=\lambda_\theta=5$ and $\lambda_w=\lambda_h=1$.
We design the anchor aspect ratio as $[0.5, 2]$ to fit to grasp poses better.
During the first 20 epochs, $\lambda_p, \lambda_r, \lambda_q$ take values of $(1, 1, 5)$ respectively. After that, they are modified to $(5, 5, 2)$ to stabilize the training.

As introduced in~\ref{subsec: data_gen}, the distance between the object and the camera is drawn from $\mathcal{U}(\sqrt{3} r_{obj}, 2 r_{obj})$. 
In reality, such condition is hard to achieve since the camera is usually fixed at a particular point. 
To tackle this, we propose to re-project the depth image into a virtual camera. 
Object's point cloud is first generated base on the depth image. $r_{obj}$ is then computed as the radiance of the point cloud's bounding ball. Next, the virtual camera's position is determined by the sampled camera-object distance. Finally, the generated point cloud is projected to the virtual camera to obtain the normalized depth image.

Fig.~\ref{fig: single_results} shows the grasp planning and grasp execution results on 12 different objects with a single stereo camera.
The top two rows of the figure show the captured RGB-Depth images. Located grasps are labeled in the depth image with red rectangles. The tilt angle $\gamma$, and the distance $z$ are neglected in the plot.
The physical grasp pose and the execution result of the planned grasp are shown in the bottom two rows. 
The algorithm is able to find robust grasps for a) small objects close to the ground, b) large objects with graspable regions, and c) objects with complex surfaces.

Table~\ref{table: compare} compares CGPN with the grasp pose detection (GPD)~\cite{gpd} algorithm, which also focuses on 6-DoF grasp planning with a single camera. 
To adopt GPD, we train a point-cloud-based grasp evaluation network with the same dataset as CGPN.
Each object is grasped three times, results in 36 trials for each algorithm.
From experiments, we observe CGPN outperforms GPD in both the grasp success rate and the computation time. One reason behind this might be the robustness of our model under the various camera angles we set during the experiment. Regarding time cost, CGPN generates valid grasps in an end-to-end manner, it does not require the sampling and evaluation procedure and therefore takes less time to plan a grasp. 

For ablation study on the data augmentation and contrastive learning module, Table~\ref{table: compare} also compares the effectiveness of adding augmentation operations on training data and using contrastive loss in the sense of object grasp success rate. As can be seen, the sim-to-real gap significantly affects the performance of the grasp proposal network. Ignoring vision noises and training on the synthetic dataset may yield poor results in practice.

\begin{table}[t]
\centering
\caption{Performance analysis of the CGPN and baselines on single object grasping tasks.
}
\label{table: compare}
\begin{tabular}{l|c|c}
\hline
\multicolumn{1}{c|}{}     & Success Rate    & Time (sec/grasp) \\ \hline\hline
GPD                         & 72.2\%          & 1.84             \\
CGPN w/o Contrastive        & 69.4\%          & 0.46             \\
CGPN w/o Data Augmentation  & 66.7\%          & 0.46             \\
CGPN                        & \textbf{75.0\%} & \textbf{0.46}    \\ 
\end{tabular}
\end{table}

\begin{figure}[h]
\begin{center}
	\includegraphics[width=2.0in]{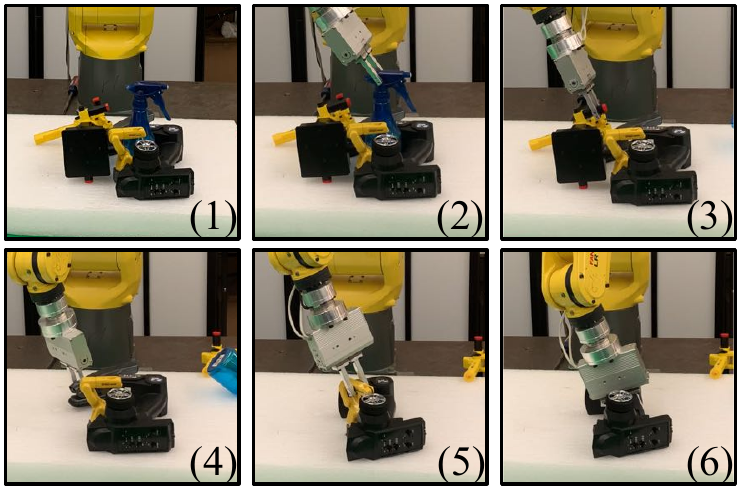}
	\caption{(1-6) show a sequence of proposed grasps in a cluttered scene.}
	\label{fig: clutter_results}
\end{center}
\end{figure}

\subsubsection*{Cluttered scene grasp}
Fig.~\ref{fig: clutter_results} shows a sequence of proposed grasps in a cluttered environment. The grasp sequence is selected according to
the graspability, or whether a collision-free grasp exists for a particular object.

\subsubsection*{Failure cases}
\begin{figure}[h]
\begin{center}
	\includegraphics[width=2.8in]{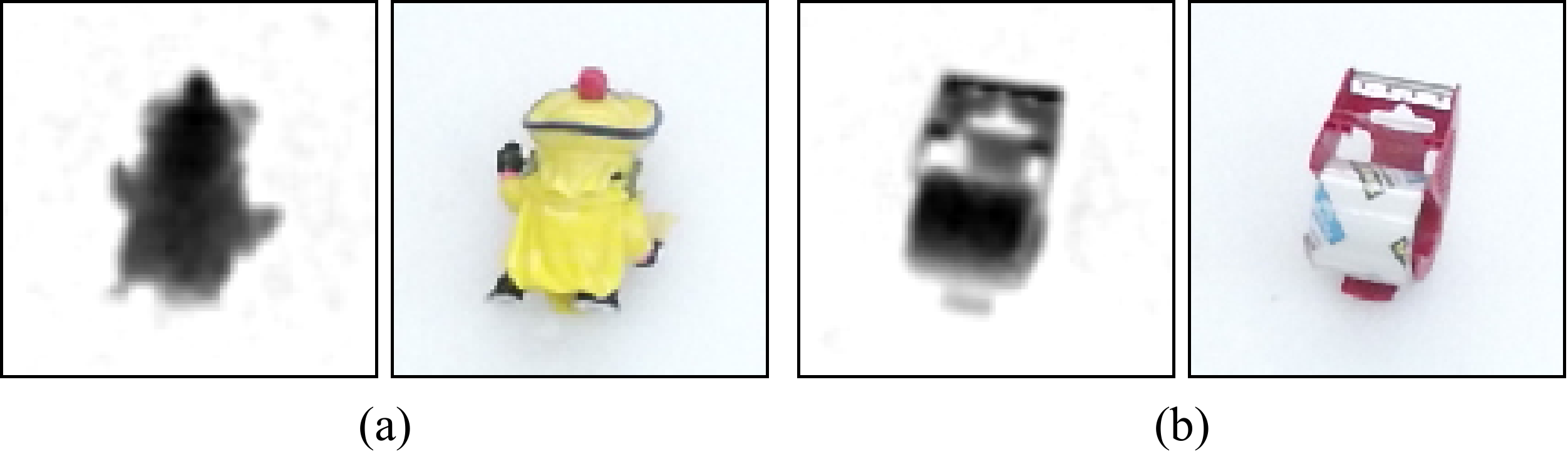}
	\caption{Two failure modes of CGPN. (a) shows the unmodeled sim-to-real gap on the object's surface, and (b) shows the limitation of the depth image in representing 6-DoF grasps.}
	\label{fig: failures}
\end{center}
\end{figure}

Fig.~\ref{fig: failures} displays two typical failures for CGPN, in which no grasp is proposed. 
The first failure mode occurs because of the unmodeled sim-to-real gap. The object surface's resolution and distance noise are not appropriately handled. 
The second type of failure occurs when no robust grasp exists with $\gamma \in [-30\degree, 30\degree]$. Compared to 3D representations, the image includes less geometric information, making the end-to-end model hard to infer 6-DoF grasp.
It appears that the performance could be improved with comprehensive data augmentations and other grasp representations.

\section{Conclusions and Future Works} 
\label{sec: conclusion}
In this paper, we present a 6-DoF contrastive grasp proposal network (CGPN) to generate robust grasps on a single-view depth images. CGPN is composed of a grasp planning module for 6-DoF grasp detection, and a contrastive learning module for sim-to-real gap reduction. 
The grasp planning module infers 6-DoF grasps base on detected robust grasp regions. 
An image encoder is used to extract the feature map, followed by a rotated region proposal network to propose planar grasps.
Feature vectors are then extracted and refined to 6-DoF grasps. 
To transfer grasp skill trained in simulation, a contrastive learning module and variant depth image processing techniques are introduced during the training.
CGPN can locate 6-DoF collision-free grasps using a single-view depth image within 0.5 second. Experiment results show that CGPN outperforms previous grasping algorithms. 
The experimental videos are available at~\cite{website}.

In the future, we would like to study the style transfer between synthetic and real depth images. 
Moreover, we believe the grasp proposal network can be designed to work with the cluttered scenes directly by modelling the influence between objects with graphs or attention mechanisms.

\addtolength{\textheight}{-1cm}   



 

\newpage
\bibliographystyle{IEEEtran}
\bibliography{references}

\end{document}